\documentclass[preprint,12pt]{elsarticle}

\usepackage{soul, color}
\usepackage{graphicx}
\usepackage{adjustbox}
\usepackage{booktabs,makecell}
\usepackage{makecell}
\usepackage{tabularx}
\usepackage{threeparttable, tablefootnote}
\usepackage{amsmath}
\newcommand{\linenumbers}{}

\usepackage{newunicodechar}

\usepackage[colorlinks,allcolors=black]{hyperref}
\usepackage{hyperref}
\hypersetup{colorlinks,allcolors=black}

\usepackage{graphicx}
\usepackage{multirow}
\usepackage[dvipsnames]{xcolor}
\usepackage{soul}
\usepackage{color}
\usepackage{url}
\usepackage{float}
\usepackage{boldline}
\usepackage{xcolor,colortbl}
\usepackage{tikz}

\usepackage{pifont}

\usepackage{multirow}
\usepackage{graphicx}
\usepackage{amssymb}
\usepackage{soul}
\usepackage[normalem]{ulem}
\usepackage{xcolor}

\usepackage{xspace}
\usepackage{tabularx}

\journal{}

\begin{document}

\begin{frontmatter}


\title{Systematic Integration of Attention Modules into CNNs for Accurate and Generalizable Medical Image Diagnosis}

\author{Zahid Ullah$^{1}$, Minki Hong $^{1}$, Tahir Mahmood $^{2}$, Jihie Kim$^{1,*}$}

\address{%
$^{1}$ \quad Department of Computer Science and Artificial Intelligence, Dongguk University, Seoul 04620, Republic of Korea \\ 
$^{2}$ \quad Division of Electronics and Electrical Engineering, Dongguk University, Seoul 04620, Republic of Korea} 

\begin{abstract}
Deep learning has become a powerful tool for medical image analysis; however, conventional Convolutional Neural Networks (CNNs) often fail to capture the fine-grained and complex features critical for accurate diagnosis. To address this limitation, we systematically integrate attention mechanisms into five widely adopted CNN architectures, namely, VGG16, ResNet18, InceptionV3, DenseNet121, and EfficientNetB5, to enhance their ability to focus on salient regions and improve discriminative performance. Specifically, each baseline model is augmented with either a Squeeze-and-Excitation block or a hybrid Convolutional Block Attention Module, allowing adaptive recalibration of channel and spatial feature representations. The proposed models are evaluated on two distinct medical imaging datasets: (1) a brain tumor MRI dataset comprising multiple tumor subtypes, and (2) a Products of Conception histopathological dataset containing four tissue categories. Experimental results demonstrate that attention-augmented CNNs consistently outperform baseline architectures across all metrics. In particular, EfficientNetB5 with hybrid attention achieves the highest overall performance, delivering substantial gains on both datasets. Beyond improved classification accuracy, attention mechanisms enhance feature localization, leading to better generalization across heterogeneous imaging modalities. This work contributes a systematic comparative framework for embedding attention modules in diverse CNN architectures and rigorously assesses their impact across multiple medical imaging tasks. The findings provide practical insights for the development of robust, interpretable, and clinically applicable deep learning–based decision support systems.

\end{abstract}

\end{frontmatter}

\linenumbers

\section{Introduction}
\label{intro}
Deep learning (DL) has revolutionized the field of medical image analysis, offering powerful tools for disease diagnosis, classification, and segmentation. Convolutional Neural Networks (CNNs), in particular, have demonstrated remarkable performance in various medical imaging tasks, owing to their ability to automatically learn discriminative features from complex image data. Specifically, the classification of phenotypes in products of Conception (PoC) and the diagnosis of brain tumors via magnetic resonance imaging (MRI) stand out as critical applications \cite{abd2021differential}.  POC, derived from spontaneous abortions, offers vital genetic information that can shed light on the causes of pregnancy loss.  Accurate phenotyping of these samples is crucial for identifying chromosomal abnormalities and genetic disorders \cite{khan2022intelligent}. Brain tumor diagnosis using MRI is similarly vital, as early and precise identification of tumor type and grade directly influences patient management and prognosis \cite{nadeem2020brain}. Leveraging pretrained CNN models such as VGG16, ResNet18, InceptionV3, DenseNet121, and EfficientNetB5 has become a common practice, as these architectures are well-established and have been trained on large-scale datasets like ImageNet, facilitating effective transfer learning for medical applications.

This study focuses on the classification of phenotypic patterns in two critical medical imaging datasets: the POC dataset, which includes samples related to spontaneous abortion, and the Brain Tumor MRI dataset. Accurate classification in these domains can provide valuable insights for early diagnosis, treatment planning, and understanding underlying pathological mechanisms. Our experimental framework is designed to explore the performance enhancements achievable by incorporating advanced attention mechanisms into standard CNN architectures. We conduct four sets of experiments to evaluate and compare model performance under varying architectural configurations. 

In the first phase, we employ the pretrained versions of VGG16, ResNet18, InceptionV3, DenseNet121, and EfficientNetB5 without any architectural modifications. These models serve as the baseline for subsequent performance comparisons.

\begin{figure*}[!ht]
     \centering
     \includegraphics[width=1\textwidth]{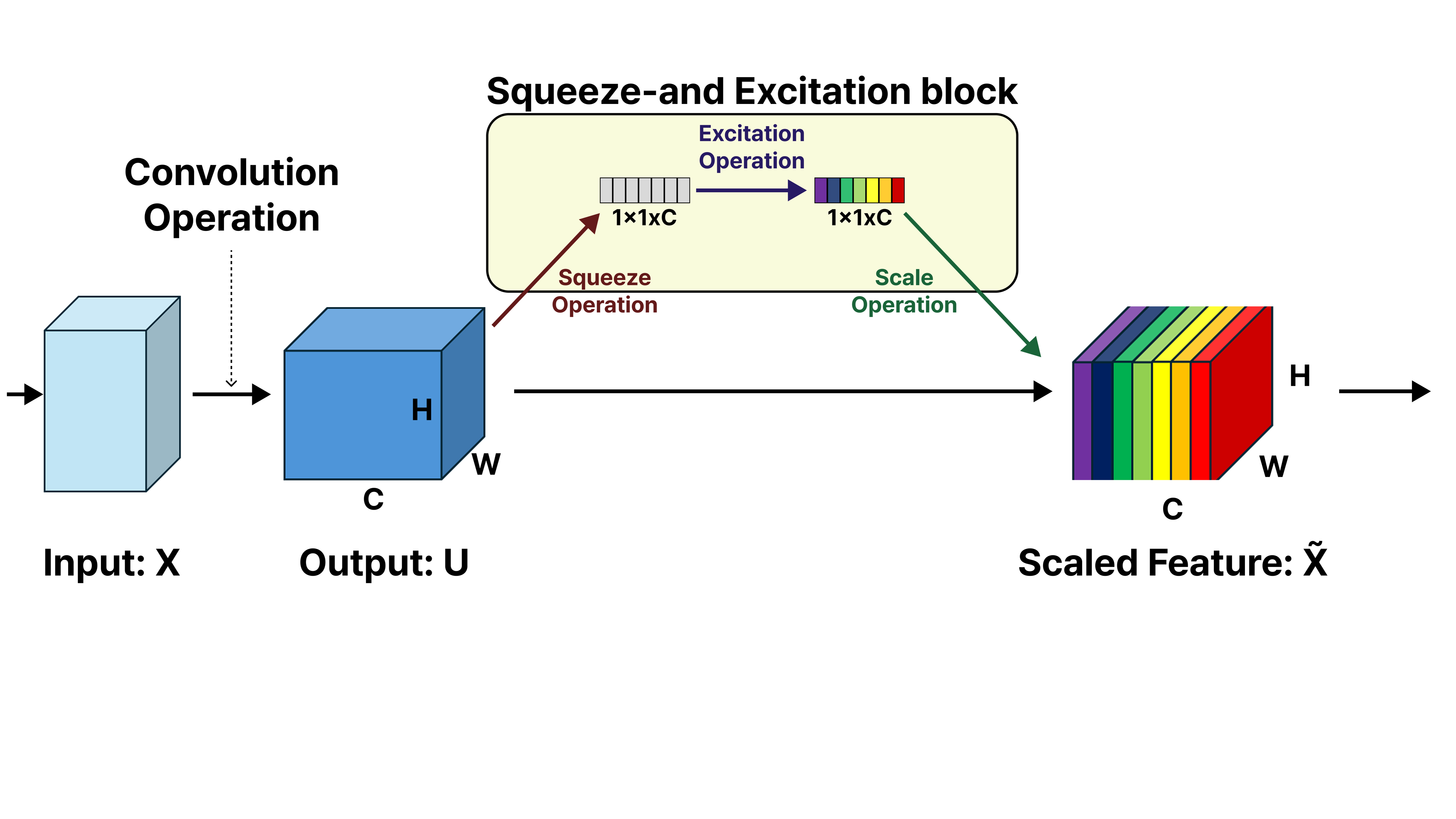}
     \caption{Structure of SE module. }
     \label{CLC}
 \end{figure*}

 In the second phase, we integrate the SE module throughout all the convolutional blocks of each pretrained model. As shown in Fig. \ref{CLC}, the SE block is introduced by Hu et al. \cite{hu2018squeeze}, which is a lightweight architectural unit designed to improve the representational power of CNNs by explicitly modeling the interdependencies between channels in feature maps. The key insight behind SE networks is that the importance of feature channels can vary depending on the input, and learning to recalibrate channel-wise responses adaptively can enhance performance in image classification and related tasks. The SE block consists of two main operations: squeeze and excitation. The squeeze operation aggregates spatial information into a channel descriptor by performing global average pooling across each feature map. This captures global context while reducing dimensionality. The excitation operation then passes this descriptor through a small two-layer fully connected neural network with a non-linearity (ReLU followed by sigmoid) to produce a set of channel-wise weights. These weights are used to rescale the original feature maps, effectively emphasizing informative channels and suppressing less useful ones.

 In the third phase, we investigate a selective integration strategy where SE modules are inserted only in deeper layers of the network. For instance, in the case of VGG16, SE blocks are added after Block 3, Block 4, and Block 5. This approach aims to focus attention refinement on higher-level, semantically richer features while minimizing computational overhead.

\begin{figure*}[!ht]
     \centering
     \includegraphics[width=1\textwidth]{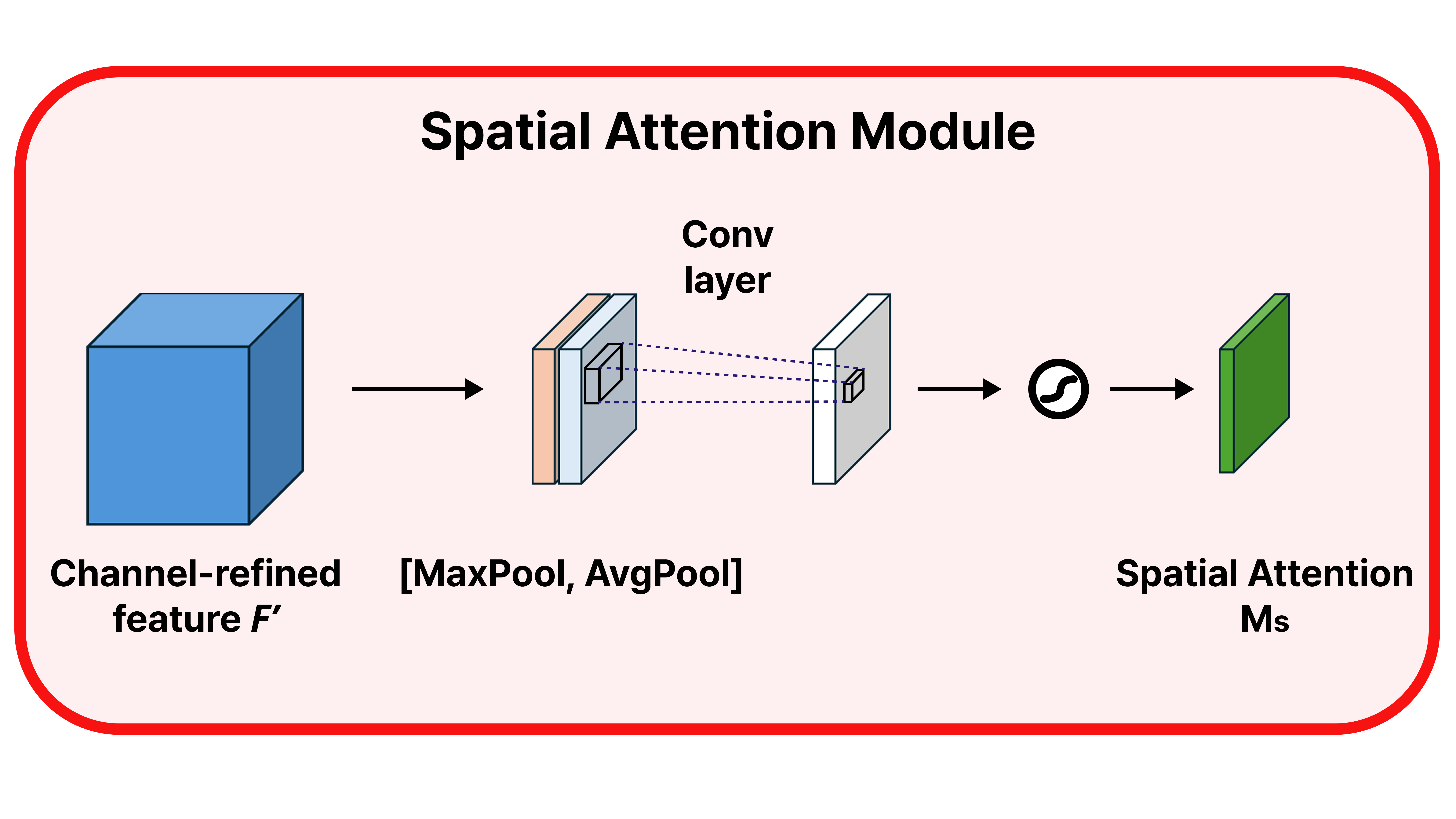}
     \caption{Structure of SA module.}
     \label{fig2}
 \end{figure*}

 In the final phase, we augment the selectively integrated SE modules with Spatial Attention (SA) mechanisms \cite{woo2018cbam}. This hybrid attention strategy allows the model to attend not only to informative channels but also to spatial regions within the feature maps, thereby enhancing localization and classification capabilities. As shown in Fig. \ref{fig2}, SA is a mechanism designed to help CNNs focus on where the most informative regions in a feature map are located. Unlike channel attention, which emphasizes what feature channels are important, SA works by assigning weights to each spatial location (i.e., pixels) across the feature map. Typically, it operates by first compressing the input feature map along the channel axis using average pooling and max pooling, producing two spatial descriptors. These are then concatenated and passed through a convolutional layer followed by a sigmoid activation to generate a SA map. This map highlights important regions and suppresses irrelevant ones, allowing the network to better capture spatial dependencies and improve performance on tasks such as image classification. 

 While recent efforts such as Improved EATFormer \cite{shisu2024improved} have explored Vision Transformer-based models with sophisticated self-attention mechanisms for medical image classification, these approaches often require substantial computational resources and large-scale datasets to achieve optimal performance. Moreover, Vision Transformer architectures typically provide results for a single novel design, limiting their generalizability across different backbone families. In contrast, our work introduces a systematic comparative framework that integrates lightweight and effective attention modules (SE and CBAM) into five widely adopted CNN architectures (VGG16, ResNet18, InceptionV3, DenseNet121, and EfficientNetB5). This design not only ensures computational efficiency and ease of deployment in resource-constrained clinical settings but also provides comprehensive insights into the impact of attention integration across diverse network families and heterogeneous imaging modalities. By addressing both performance and practicality, our study complements Transformer-based approaches while offering broader applicability and clinical relevance.

Specifically, the primary objective of this study is to systematically evaluate how different attention mechanisms, such as global, selective, and hybrid, affect the classification performance of pretrained CNN models in challenging medical image classification tasks. Our experimental results reveal that incorporating attention modules, particularly in a selective and spatially guided manner, can significantly improve the discriminative power of CNNs without incurring excessive computational costs. The contributions of this study are stated below:

\begin{itemize}
    \item We systematically integrate lightweight attention modules (SE and CBAM) into five widely adopted CNN architectures, creating multiple architectural variants to assess the impact of channel and spatial-attention mechanisms.

\item  Unlike prior works that focus on a single backbone or Transformer design, our study provides a unified framework that compares attention integration strategies across diverse CNN families, offering practical design insights for medical image classification.

\item We rigorously validate the proposed models on two distinct medical imaging datasets, brain tumor MRI (multi-subtype) and Products of Conception histopathology (multi-class tissue categories), demonstrating robustness across radiology and pathology domains.

\item  Experimental results show that attention-augmented CNNs consistently outperform their baseline counterparts, with EfficientNetB5 + hybrid attention achieving the highest accuracy. Moreover, attention integration improves feature localization, enhancing interpretability and clinical relevance.
\end{itemize}

The remaining paper is organized as follows: Section \ref{related} presents a comprehensive review of background knowledge to contextualize our research. Next, in Section \ref{pm} we detail our proposed methodology, including the dataset. Section \ref{Results} consists of the results and discussion. We then describe limitations and future work in section \ref{limitation}. Finally, in section \ref{conclusion}, we present the conclusions.

\section{Related work}\label{related}
DL techniques, particularly CNNs have revolutionized medical image analysis by automating the processing and diagnostics of various medical conditions \cite{chung2025artificial}. CNNs have shown great success in medical image analysis, allowing for great progress in computer-aided diagnosis \cite{yu2021convolutional}. CNN models can automatically extract relevant characteristics from brain MRI, increasing cancer detection sensitivity and specificity compared to traditional computer-aided detection systems \cite{ullah2025hierarchical,ullah2025hybrid}. CNN-based frameworks have demonstrated promising results in improving diagnostic accuracy through the multi-class classification of neurodegenerative diseases \cite{asif2025brain}. The application of CNNs in brain tumor detection has also been extensively explored, with studies demonstrating high accuracy in identifying and classifying different types and grades of tumors \cite{babayomi2023convolutional,srinivasan2024hybrid}. Transfer learning \cite{raffel2020exploring}, which leverages pre-trained models on large datasets like ImageNet, has proven to be a valuable strategy in medical imaging due to the limited size of medical datasets \cite{kim2022transfer}. Pre-trained CNN models, such as those utilized in our study, offer a strong starting point for medical image analysis tasks by leveraging the knowledge gained from extensive training on large datasets \cite{sun2024brain}. Fine-tuning these pre-trained models on specific medical datasets can lead to significant improvements in performance compared to training models from scratch. To further enhance the capabilities of these pre-trained models, attention mechanisms, such as SE modules and SA, have been incorporated to allow the network to focus on the most salient features in the images \cite{thakur2024deep}. Attention mechanisms have enabled models to focus on relevant regions of interest within an image. 

To further extend the existing literature on SE networks, Rongjun et al. \cite{ge2021convolutional} propose a novel approach that integrates SE blocks into a 1D residual CNNs specifically designed for ECG arrhythmia detection. The framework leverages temporal 1D convolutions combined with SE modules to adaptively highlight arrhythmia‑relevant channel features while suppressing redundant information, eliminating the need for preprocessing steps such as denoising. The incorporation of residual connections facilitates more stable training and efficient optimization. Experimental validation on the CPSC 2018 12‑lead ECG dataset and the PhysioNet/CinC 2017 dataset demonstrates that the proposed architecture achieves high classification accuracy across multiple arrhythmia types, underscoring its potential for reliable and efficient automated ECG diagnosis.

Li et al. \cite{li2020epileptic} proposed a unified temporal spectral SE architecture that jointly captures multi‑scale temporal dynamics and multi‑level spectral features from raw EEG data. The network integrates temporal convolutional blocks to model nonstationary temporal patterns alongside spectral convolution blocks capturing frequency‑band characteristics, and then fuses them via a novel SE module to highlight the most discriminative channel-wise features. To address overfitting due to scarce seizure events, the model is trained with an information‑maximizing loss based on maximum mean discrepancy. Evaluated on three public EEG datasets, such as the Bonn dataset, the CHB‑MIT dataset, and the TUH Seizure Corpus, CE‑stSENet achieves competitive detection performance compared to state‑of‑the‑art methods, demonstrating the effectiveness of jointly modeling temporal and spectral domains in seizure recognition.

Kitada et al. \cite{kitada2018skin} fine-tuned an ensemble of pre-trained SENet using a mean‑teacher semi‑supervised framework, combining labeled and unlabeled skin lesion images to bolster effectiveness. They augmented training with dermatology‑focused transformations, including random cropping, body‑hair augmentation, rotation, flips, random erasing, and between-class mixing to enrich variability and better mimic real-world lesion appearances. A five‑fold cross‑validation ensemble of SE‑ResNet‑101 models was employed, and test‑time augmentation was used to aggregate predictions. The combined strategy significantly improved balanced accuracy from 79.2\% without custom augmentation to 87.2\% with body‑hair augmentation on the official ISIC 2018 validation set, demonstrating the value of SE‑based deep architectures and targeted semi‑supervised training strategies in medical image classification.

 While prior research has demonstrated the effectiveness of Squeeze-and-Excitation (SE) networks and attention mechanisms in specific domains such as ECG analysis \cite{ge2021convolutional}, EEG-based seizure detection \cite{li2020epileptic}, and dermatological lesion classification \cite{kitada2018skin}, these studies are often task-specific and limited to a single imaging modality. In contrast, our work systematically investigates the integration of attention mechanisms into multiple widely adopted CNN architectures, VGG16, ResNet18, InceptionV3, DenseNet121, and EfficientNetB5 across two heterogeneous medical imaging modalities, brain tumor MRI and histopathological images of POC. Unlike prior works that focus narrowly on one type of signal or dataset, our comparative framework evaluates both channel-focused (SE) and hybrid channel-spatial (CBAM) attention modules, providing a broader understanding of their impact on feature recalibration and localization. Our findings demonstrate that attention-augmented CNNs not only enhance classification accuracy but also achieve superior generalization across diverse datasets, with EfficientNetB5 + CBAM achieving the highest performance. This systematic and modality-agnostic evaluation distinguishes our study from earlier works and contributes practical insights for the design of robust and generalizable attention-based models for clinical decision support.
 
\section{Materials and Methods} 
\label{pm}
The methodology employed in this study was structured around a series of experiments designed to evaluate the performance of pre-trained CNN models and their enhanced variants in the context of classifying phenotypes in POC and brain tumor MRI data. 

We first established baseline results by directly applying the pre-trained CNN models (VGG16 \cite{simonyan2014very}, ResNet18 \cite{he2016deep}, InceptionV3 \cite{szegedy2015going}, EfficientNetB5 \cite{tan2019rethinking}, and DenseNet121 \cite{huang2017densely}) to our datasets. This step involved fine-tuning the pre-trained models (VGG16, ResNet18, InceptionV3, EfficientNetB5, and DenseNet121) on the specific classification tasks without any modifications to their architectures. This enabled us to gauge the inherent capabilities of each model on our tasks, providing a benchmark against which subsequent enhancements could be measured.

In the second phase, we augmented the pre-trained models with SE modules. SE modules are designed to improve the representational capacity of CNNs by adaptively recalibrating channel-wise feature responses. This allows the model to emphasize informative features and suppress less relevant ones.

Recognizing that the indiscriminate application of SE modules may not always yield optimal results, we explored a more selective approach in the third experiment. In VGG16, we strategically placed SE modules after specific convolutional blocks (block 3, block 4, and block 5), guided by the intuition that these blocks capture progressively more complex features. 

In the final experiment, we incorporated SA mechanisms alongside the selective SE modules. SA enables the model to selectively focus on the most relevant spatial regions within an image, complementing the channel-wise recalibration provided by the SE modules. By combining these two types of attention mechanisms, our intention was to create a model that is both sensitive to feature interdependencies across channels and spatially aware of the most informative regions in the image. 

\subsection{Datasets}
\begin{figure*}[!ht]
     \centering
     \includegraphics[height=.65\textwidth]{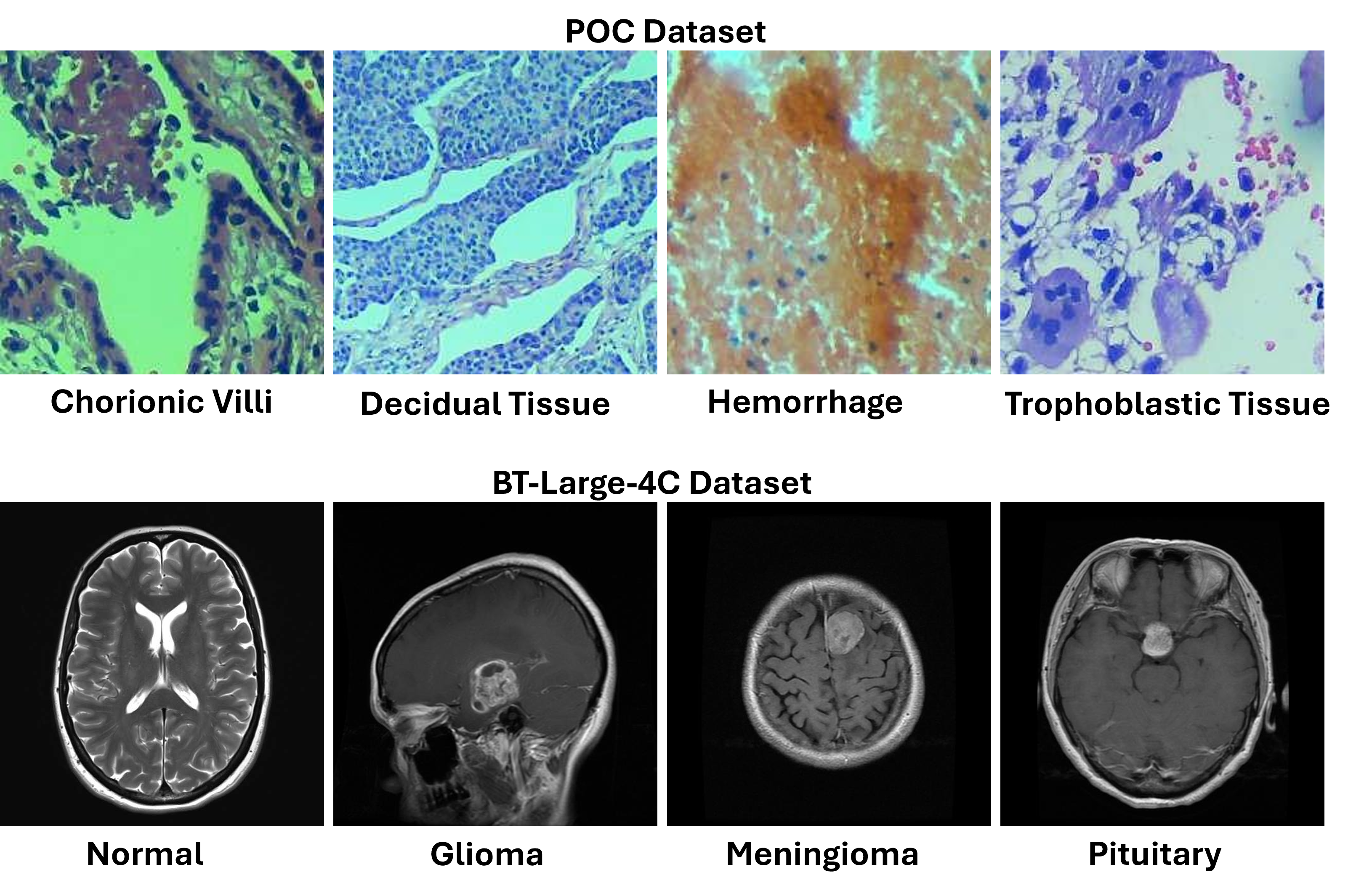}
     \caption{Representative samples from the POC and BT-Large-4C datasets.}
     \label{dataset}
 \end{figure*}

In this study, we employed two publicly available datasets, namely, the POC dataset and the BT-Large-4C dataset as shown in Fig. \ref{dataset}. A summary of these datasets is presented in Table \ref{POC_BT}, while detailed descriptions of each dataset are provided in the following subsections.

\begin{table}[!ht]
\centering
\caption{Details of the datasets.}
\scalebox{1.2}{
\begin{tabular}{cccc}
\hline
\textbf{Dataset}      & \textbf{Number of Class} & \textbf{Training Set} & \textbf{Test Set} \\ \hline
Product of Conception & 4                        & 4155                  & 1511             \\
BT-large-4c           & 4                        & 2611                  & 653               \\ \hline
\end{tabular}
\label{POC_BT}
}
\end{table}

\subsubsection{Products of Conception Dataset}
We utilized the POC dataset \cite{mahmood2024computer}, which is publicly available for research purposes. The dataset consists of histopathological image samples representing four different tissue categories, namely, chorionic villi, decidual tissue, hemorrhage, and trophoblastic tissue. The training set contains a total of 4,155 samples, distributed as follows: 1,391 chorionic villi, 926 decidual, 1,138 hemorrhage, and 700 trophoblastic tissue images. The testing set includes 1,511 samples, with 390 chorionic villi, 349 decidual, 421 hemorrhage, and 351 trophoblastic tissue images.

\subsubsection{Brain Tumor Dataset}
We have also employed a publicly available brain tumor MRI dataset acquired from the Kaggle repository \footnote{https://www.kaggle.com/sartajbhuvaji/brain-tumor-classification-mri}. The dataset comprises 3,064 T1-weighted contrast-enhanced brain MR images, covering three common tumor types: gliomas, meningiomas, and pituitary tumors. To facilitate a more comprehensive evaluation, we extended the dataset into four categories by additionally including normal brain MR images, and we refer to this dataset as BT-Large-4c. The BT-Large-4c dataset thus consists of four distinct classes, namely, Normal, Glioma tumor, Meningioma tumor, Pituitary tumor. This dataset provides a sufficiently large and diverse collection of images for training and validating DL models, ensuring robust evaluation of tumor classification performance across different tumor types.

To address the limited size of our MRI dataset, we employed image augmentation, a technique that artificially increases the entries in a dataset by altering the original images. This approach creates multiple variations of each image by adjusting parameters such as scale, rotation, position, and brightness, among other attributes. Studies have shown \cite{perez2017effectiveness,yang2022image} that augmenting datasets can enhance model classification accuracy more effectively than collecting additional data.

For our image augmentation process, we applied two specific methods, rotation and horizontal flipping. The rotation involved randomly rotating the input images by 90 degrees one or more times. Afterward, horizontal flipping was performed on each rotated image, further enriching the dataset with additional training samples.

\subsection{Baseline CNN Architectures}

\subsubsection{VGG16}
VGG16 is a deep CNN proposed by Simonyan and Zisserman \cite{simonyan2014very}, which achieved top performance in the ImageNet Large-Scale Visual Recognition Challenge (ILSVRC) 2014. The architecture is characterized by its uniform use of small $3 \times 3$ convolutional filters, applied with a stride of 1 and padding to preserve spatial resolution. These convolutional layers are stacked sequentially, enabling the network to learn increasingly abstract hierarchical features while keeping the number of parameters manageable compared to larger filters. The network consists of 13 convolutional layers and 3 fully connected layers, totaling 16 weight layers, hence the name VGG16.

\begin{figure*}[!ht]
     \centering
     \includegraphics[height=55mm]{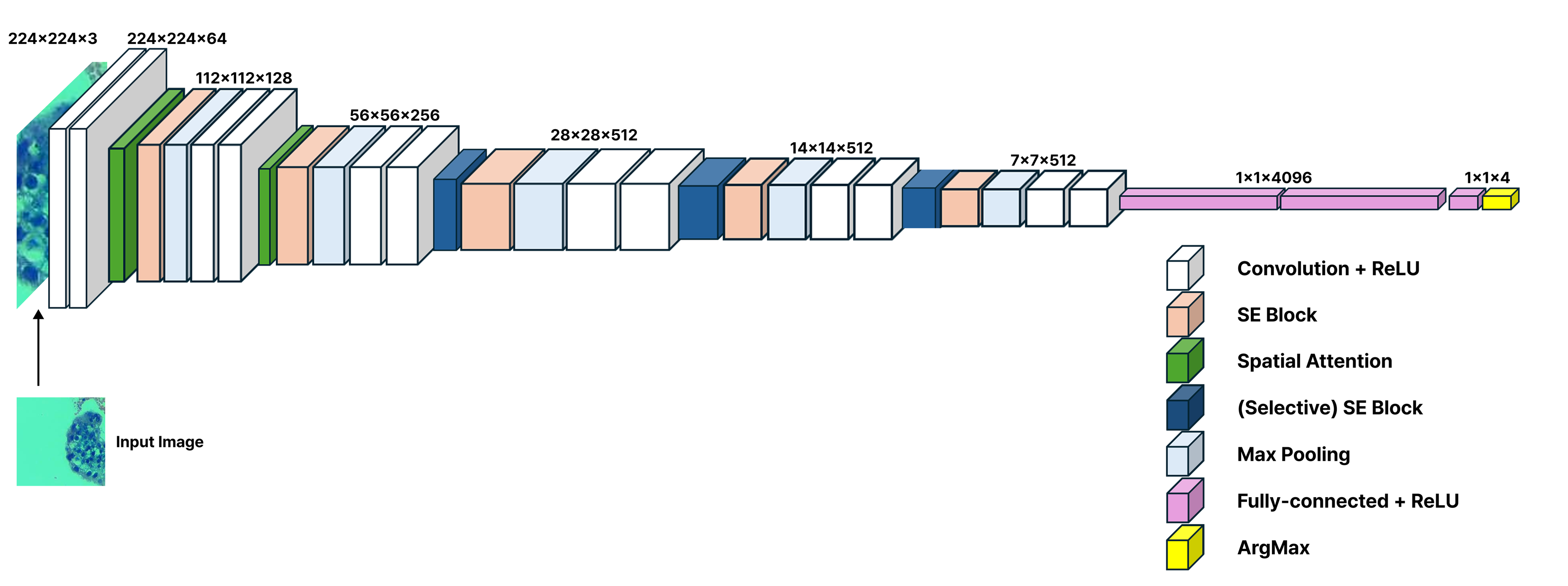}
     \caption{Architectural variants of VGG16 with integrated attention mechanisms. In the first variant (VGG16-SE v1), SE blocks are inserted after every convolutional layer to recalibrate channel responses. The second variant (VGG16-SE v2) applies SE blocks selectively, placing them only after the last convolution of blocks 3, 4, and 5 to emphasize higher-level features with reduced complexity. The third variant (VGG16-SE-SA) combines channel and SA by inserting SA (SA) modules after the last convolutions of blocks 1 and 2, and SE blocks after the last convolutions of blocks 3, 4, and 5, thereby enhancing both spatial and channel feature representations.}
     \label{fig3}
 \end{figure*}

The convolutional layers are grouped into five blocks, each followed by a max-pooling layer of size $2 \times 2$ with a stride of 2, progressively reducing spatial dimensions while retaining important feature information. After the final convolutional block, the extracted feature maps are flattened and passed through three fully connected layers, with the last one typically using a softmax activation for classification. Despite its depth, the simplicity and uniformity of the architecture make VGG16 widely adopted as a baseline in many computer vision tasks, including medical image analysis, due to its strong feature extraction capability and availability of pre-trained weights on large-scale datasets such as ImageNet.

To explore the role of attention in convolutional networks, we constructed three VGG16-based variants with integrated attention mechanisms, as shown in Fig. \ref{fig3}. 

The first variant, VGG16-SE v1, augments the baseline by inserting SE blocks after every convolutional layer. This design enables channel-wise recalibration at all depths of the network, ensuring that informative channels are emphasized and redundant ones are suppressed throughout the feature extraction hierarchy. 

The second variant, VGG16-SE v2, adopts a more selective strategy. SE blocks are applied only after the last convolutional layer of blocks 3, 4, and 5, focusing the recalibration process on higher-level semantic features. This reduces additional computational overhead while still improving representational power in the deeper layers, where abstraction is most critical. 

The third variant, VGG16-SE-SA, incorporates both channel and spatial-attention mechanisms. Specifically, SA modules are introduced after the last convolution of blocks 1 and 2 to highlight salient spatial regions, while SE blocks are integrated after the last convolutions of blocks 3, 4, and 5 to refine channel responses. This hybrid design leverages the complementary strengths of spatial and channel attention, allowing the network to enhance both where and what to focus on within feature maps.

\subsubsection{ResNet18}
ResNet18 is a residual CNN introduced by He et al. \cite{he2016deep}, which marked a breakthrough in DL by addressing the vanishing gradient problem through the concept of residual learning. Unlike traditional deep neural networks that directly learn mappings from input to output, ResNet introduces shortcut connections (or skip connections) that allow the network to learn residual functions instead of unreferenced mappings. This enables the effective training of deeper architectures without suffering from performance degradation.

\begin{figure*}[!ht]
     \centering
     \includegraphics[width=1\textwidth]{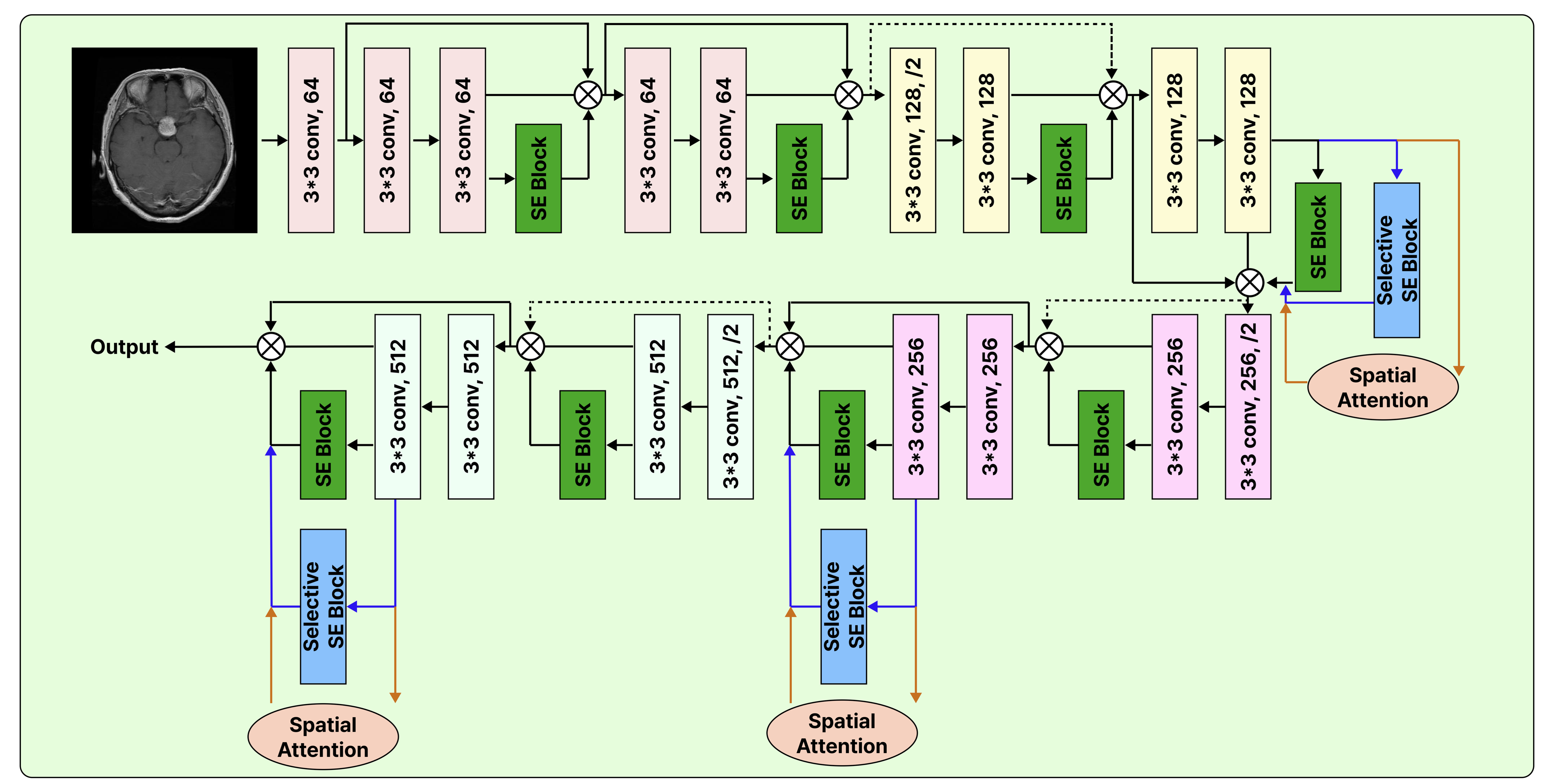}
     \caption{Architectural variants of ResNet18 with integrated attention mechanisms. The first variant (SEResNet18 v1) replaces BasicBlocks with SE-enhanced blocks for channel recalibration. The second variant (SEResNet18 v2) inserts SE blocks after layers 2, 3, and 4 to emphasize mid- and high-level features. The third variant (SEResNet18-SA) combines SE and SA by applying SE and SA modules sequentially after layers 2, 3, and 4, enabling joint channel- and spatial-wise feature refinement.}
     \label{fig4}
 \end{figure*}

The ResNet18 architecture consists of 18 learnable layers, including 17 convolutional layers and one fully connected layer. The network is organized into five stages: an initial $7 \times 7$ convolutional layer with a stride of 2, followed by a $3 \times 3$ max-pooling layer, and then four residual blocks. Each block is composed of two $3 \times 3$ convolutional layers with batch normalization and ReLU activation, accompanied by an identity shortcut connection that bypasses one or more layers. These shortcuts can be identity mappings when the input and output dimensions match, or $1 \times 1$ convolutions when adjustment of dimensions is required.

By leveraging residual connections, ResNet18 is able to extract more discriminative features while maintaining computational efficiency, making it suitable for applications where both accuracy and reduced complexity are essential. Furthermore, its pre-trained weights on large-scale datasets such as ImageNet have made ResNet18 a popular backbone for various computer vision tasks, including classification, segmentation, and medical image analysis.

As shown in Fig. \ref{fig4}, we also explore attention-enhanced variants of ResNet18 to assess how residual connections interact with SE and SA modules. Three variants were constructed. The first, SEResNet18 v1, extends the original ResNet18 by replacing each BasicBlock with its SE-enhanced counterpart. This allows channel recalibration to occur within every residual unit, ensuring that feature maps are adaptively reweighted across all layers of the network. 

The second, SEResNet18 v2, adopts a more selective strategy by inserting SE blocks only after layers 2, 3, and 4. This design emphasizes mid and high-level features, where semantic abstraction becomes more prominent, while also reducing computational complexity compared to uniformly applying SE across all residual blocks. 

The third variant, SEResNet18-SA, combines both channel and spatial-attention mechanisms. Here, SE and SA modules are applied sequentially after layers 2, 3, and 4. The SE modules adaptively recalibrate channel responses, while the SA modules capture spatial correlations within feature maps. By integrating both forms of attention, this hybrid variant enhances the model’s ability to focus on the most informative channels and spatial regions simultaneously.

\subsubsection{InceptionV3}
InceptionV3, proposed by Szegedy et al. \cite{szegedy2016rethinking}, is a deep CNN designed to achieve high accuracy with efficient computational cost. It is a refined version of the original Inception architecture, focusing on factorization techniques, dimensionality reduction, and optimized computational efficiency. The key idea of the Inception module is to capture multi-scale spatial information by applying multiple convolutional filters of different sizes ($1 \times 1$, $3 \times 3$, and $5 \times 5$) in parallel within the same layer, along with pooling operations. Their outputs are concatenated, allowing the network to extract both fine- and coarse-grained features simultaneously.

\begin{figure*}[!ht]
     \centering
     \includegraphics[width=1\textwidth]{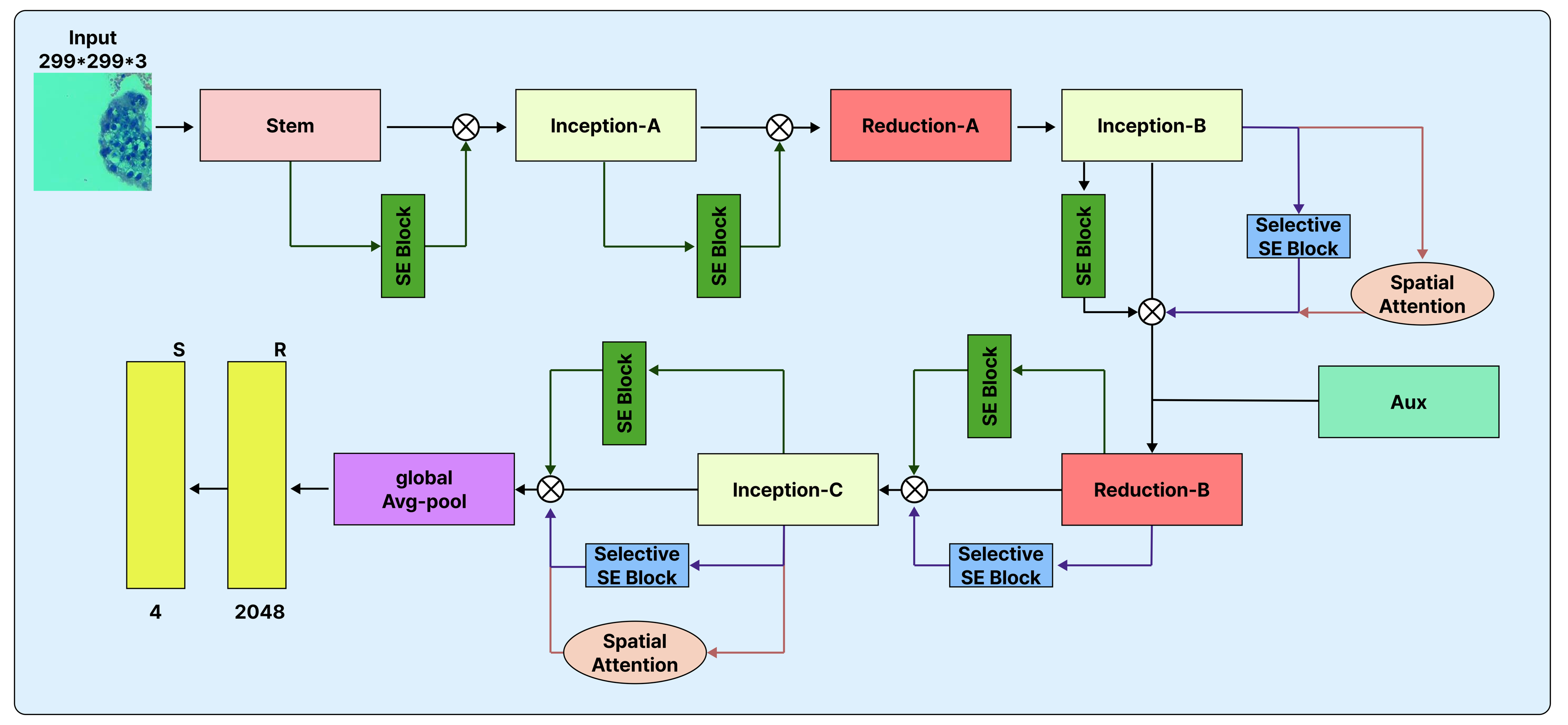}
     \caption{Architectural variants of InceptionV3 with integrated attention mechanisms. The first variant (InceptionV3-SE v1) introduces SE blocks after selected stages to recalibrate channel responses. The second variant (InceptionV3-SE v2) applies SE blocks specifically after Inception-C, Inception-D, and Inception-E modules for targeted channel-wise refinement. The third variant (InceptionV3-SE-SA) combines SE and SA (SA), adding SA after Inception-B and before global pooling, alongside SE after Inception-C, D, and E, to jointly enhance spatial and channel feature representations.}
     \label{fig5}
 \end{figure*}

InceptionV3 introduces several architectural improvements over earlier versions. Notably, it replaces larger convolutions (e.g., $5 \times 5$) with two consecutive $3 \times 3$ convolutions, significantly reducing the number of parameters without sacrificing representational power. Additionally, asymmetric convolutions (e.g., $1 \times 3$ followed by $3 \times 1$) are employed to further reduce computational complexity. Batch normalization is extensively used to stabilize and accelerate training, while auxiliary classifiers are added during intermediate layers to provide additional gradient flow and reduce overfitting.

The final architecture consists of multiple stacked Inception modules followed by a global average pooling layer and fully connected layers, with a softmax classifier for output. Due to its balance between accuracy and computational efficiency, InceptionV3 has been widely adopted in a range of image classification and medical imaging tasks, especially when resource constraints are a concern.

InceptionV3 provides a flexible architecture composed of stacked Inception modules, making it a suitable candidate for exploring the integration of attention mechanisms. To investigate their effect, three attention-enhanced variants of InceptionV3 were developed as shown in Fig. \ref{fig5}.  

The first variant, InceptionV3-SE v1, incorporates SE blocks after selected stages of the network. This placement recalibrates channel responses in intermediate feature maps, allowing the model to adaptively emphasize the most informative filters while suppressing irrelevant ones.  

The second variant, InceptionV3-SE v2, applies SE blocks more strategically, inserting them only after the Inception-C, Inception-D, and Inception-E modules. These modules correspond to deeper parts of the network where semantic abstraction is strongest. By concentrating SE recalibration at these stages, the model benefits from enhanced high-level feature representations while avoiding excessive complexity.  

The third variant, InceptionV3-SE-SA, integrates both channel and spatial-attention mechanisms. In this design, SA modules are introduced after the Inception-B module and again before the global pooling layer, ensuring that salient spatial regions are highlighted at both mid-level and final feature stages. Meanwhile, SE blocks are applied after Inception C, D, and E to refine channel responses. This hybrid configuration leverages the complementary strengths of spatial and channel attention, resulting in more robust and discriminative feature learning.

\subsubsection{EfficientNetB5}
EfficientNetB5, introduced by Tan and Le \cite{tan2019rethinking}, is part of the EfficientNet family of CNNs designed through a principled compound scaling method. Unlike traditional approaches that scale a baseline network by arbitrarily increasing depth, width, or input resolution, EfficientNet employs a compound coefficient to uniformly scale all three dimensions in a balanced manner. This approach results in models that achieve state‑of‑the‑art accuracy while being computationally efficient.

\begin{table}[!ht]
\centering
\caption{Positions of the SE and SA modules integrated into different variants of the EfficientNet-B5 architecture, indicating their insertion after specific blocks within the feature extraction pipeline.}
\scalebox{0.6}{
\begin{tabular}{lll}
\hline
\multicolumn{1}{c}{\textbf{Model}}   & \multicolumn{1}{c}{\textbf{SE Placement}}                                                                                                                                                   & \multicolumn{1}{c}{\textbf{SA Placement}} \\ \hline
\textbf{Variant 1 (MBConv + SE)}     & \begin{tabular}[c]{@{}l@{}}Inside the custom MBConv block (applied after the \\ depthwise convolution and before the projection layer)\end{tabular}                                                                  & None                                                          \\
\textbf{Variant 2 (SE after blocks)} & \begin{tabular}[c]{@{}l@{}}- After Block 2 (end of index 4, output channels = 176)\\ - After Block 3 (end of index 9, output channels = 512)\\ - After Block 4 (end of index 13, output channels = 512)\end{tabular} & None                                                          \\
\textbf{Variant 3 (SE + SA)}         & \begin{tabular}[c]{@{}l@{}}- After Block 2 (index 4)- After Block 3 (index 9)\\ - After Block 4 (index 13)\end{tabular}                                                                                              & - After Block 3 (index 9, right after SE3)                    \\ \hline
\end{tabular}
}
\label{efficientnet}
\end{table}

The EfficientNet architecture is built on a baseline network optimized through neural architecture search (NAS), and it primarily leverages Mobile Inverted Bottleneck Convolution (MBConv) blocks with SE modules. The MBConv blocks expand the input channels, apply depthwise separable convolutions, and then project the channels back to a smaller dimension, which significantly reduces the number of parameters and floating‑point operations (FLOPs). The integrated SE modules adaptively recalibrate channel‑wise feature responses, allowing the network to focus on more informative features.

EfficientNetB5 scales the baseline EfficientNet architecture with a compound coefficient that increases the network’s depth, width, and resolution beyond smaller variants (e.g., B0–B4), striking a balance between accuracy and efficiency. The model accepts larger input resolutions (456 × 456) compared to lower‑order variants, enabling it to capture finer details in images. The final layers include global average pooling followed by FC layers with a softmax classifier.

Due to its efficient design and strong representational capability, EfficientNetB5 has become a popular backbone for challenging computer vision tasks, including medical image classification and segmentation, where high accuracy and computational efficiency are equally critical.

To further enhance its representational power, we developed three attention-augmented variants of EfficientNet-B5, as summarized in Table~\ref{efficientnet}. Each variant integrates SE and SA modules at specific positions in the feature extraction pipeline. 

The first variant, EfficientNet-B5 (MBConv + SE), modifies the original MBConv blocks by incorporating SE modules inside each block. Specifically, the SE operation is applied after the depthwise convolution and before the projection layer. This design ensures that channel responses are recalibrated at every stage of the network, providing fine-grained channel-wise attention.  

The second variant, EfficientNet-B5 (SE after blocks), introduces SE blocks selectively after key stages of the architecture, namely after Block 2 (index 4, output channels = 176), Block 3 (index 9, output channels = 512), and Block 4 (index 13, output channels = 512). This selective placement emphasizes mid and high-level feature representations, improving semantic feature quality while reducing additional computational cost compared to applying SE uniformly across all MBConv blocks.  

The third variant, EfficientNet-B5 (SE + SA), combines both channel and spatial attention. In this design, SE blocks are applied after Block 2, Block 3, and Block 4, while an SA module is added after Block 3, directly following SE3. This hybrid integration enables complementary refinement, where SE modules enhance channel selectivity and SA modules highlight salient spatial regions. Together, they provide a more comprehensive attention mechanism that strengthens both ``what" and ``where" aspects of feature learning.

\subsubsection{DenseNet121}
DenseNet121, proposed by Huang et al. \cite{huang2017densely}, is a CNN that introduces dense connectivity between layers to improve feature reuse and gradient flow. In DenseNet, each layer receives inputs from all preceding layers and passes its own feature maps to all subsequent layers within the same block. This dense connectivity pattern ensures that information and gradients flow more effectively throughout the network, alleviating the vanishing gradient problem and improving parameter efficiency.

\begin{figure*}[!ht]
     \centering
     \includegraphics[width=1\textwidth]{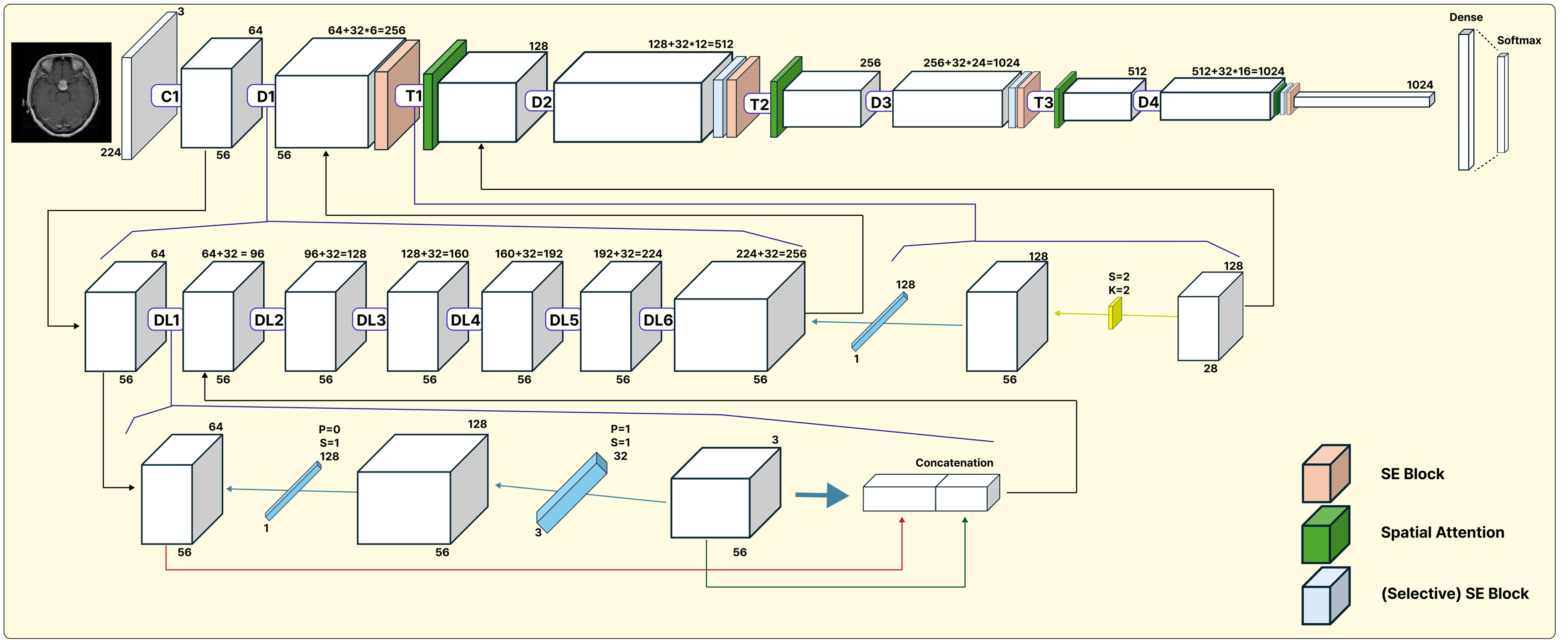}
     \caption{Architectural variants of DenseNet121 with integrated attention mechanisms. The first variant (DenseNet121-SE v1) applies SE blocks after each denseblock (1–4) to recalibrate channel responses. The second variant (DenseNet121-SE v2) selectively inserts SE blocks after denseblocks 2, 3, and 4 for focused channel-wise refinement. The third variant (DenseNet121-SE-SA) combines SE and SA (SA), placing SE after denseblocks 2–4 and SA after transitions 1–3 and before global average pooling, enabling joint channel and spatial feature enhancement.}
     \label{fig6}
 \end{figure*}

The architecture of DenseNet121 consists of 121 layers, organized into four dense blocks separated by transition layers. Each dense block contains multiple convolutional layers, where the feature maps are concatenated (not summed) with those from previous layers, resulting in highly diversified feature representations. Transition layers, which include $1 \times 1$ convolutions followed by $2 \times 2$ average pooling, are placed between dense blocks to control the growth of feature maps and reduce computational cost. The network begins with a $7 \times 7$ convolution and max‑pooling, and it concludes with a global average pooling layer and a fully connected softmax classifier.

A key advantage of DenseNet121 is its parameter efficiency. Because features are reused across layers, DenseNet requires fewer parameters compared to architectures of similar depth, while achieving high accuracy. This makes DenseNet121 particularly well‑suited for tasks with limited training data, such as medical image analysis, where both effective feature extraction and reduced risk of overfitting are critical.

To investigate the role of attention mechanisms, we developed three DenseNet121 variants with integrated SE and SA modules as shown in Fig. \ref{fig6}.  

The first variant, DenseNet121-SE v1, enhances the baseline by applying SE blocks after each of the four dense blocks. This design ensures consistent channel recalibration across all hierarchical feature levels, from low to high-level representations.  

The second variant, DenseNet121-SE v2, takes a more selective approach by inserting SE blocks only after dense blocks 2, 3, and 4. Since deeper dense blocks extract more abstract and semantically rich features, this strategy focuses computational resources on recalibrating higher-level representations, thereby balancing performance gains with efficiency.  

The third variant, DenseNet121-SE-SA, integrates both channel and spatial attention. SE blocks are applied after dense blocks 2–4 to refine channel responses, while SA modules are inserted after transitions 1–3 and before global average pooling to highlight spatially salient regions at multiple stages of the network. This hybrid configuration enables complementary attention, where SE emphasizes ``what” features to prioritize and SA captures ``where” in the spatial domain those features are most important.

\subsection{Experimental Framework}
\subsubsection{Baseline Fine-Tuning of Pre-trained CNNs}
In the first phase of our study, we establish baseline models using well-known CNNs, namely VGG16, ResNet18, InceptionV3, DenseNet121, and EfficientNetB5. For consistency and to leverage prior knowledge, we utilize their pretrained versions on the ImageNet dataset. These models are adopted without any architectural modifications, ensuring that their standard configurations serve as a fair and reliable benchmark.

Fine-tuning is performed by replacing the final classification layer with a new fully connected layer tailored to the number of target classes in our dataset. The earlier convolutional layers are initialized with pretrained weights to transfer generic feature representations, while the final layers are trained from scratch to adapt to the specific task at hand. 

This baseline fine-tuning approach enables a direct comparison of model performance across architectures, providing a reference point for evaluating subsequent methodological enhancements.

\subsubsection{Integration of SE Modules}
To enhance the representational capacity of the baseline networks, we integrated SE modules into their convolutional blocks. 
The SE block, as shown in Fig. \ref{CLC}, performs channel-wise attention by adaptively recalibrating feature maps. It achieves this by first applying global average pooling to squeeze spatial information into a channel descriptor, followed by two fully connected layers (implemented via $1\times1$ convolutions) with a reduction ratio of 16, and finally a sigmoid activation to generate channel-wise weights. These weights are multiplied by the original feature maps, thus emphasizing informative channels while suppressing less useful ones.

For instance, in our modified VGG16 architecture, SE blocks are placed after each convolutional layer, and immediately before the subsequent pooling operation (if present). Specifically, the SE modules are integrated after every convolutional layer in all five convolutional blocks of VGG16. For example, in the first block, each of the two convolutional layers is followed by an SE module, and then by a max-pooling layer. The same pattern is repeated in the subsequent blocks, ensuring that channel attention is applied throughout the feature extraction hierarchy. This design allows the model to dynamically adapt channel dependencies at multiple depths of abstraction. The classifier part of VGG16 is retained from the pretrained model, except that the final fully connected (FC) layer is replaced to match the number of target classes (i.e., four in our case). The overall integration ensures that the benefits of ImageNet-pretrained feature extraction are preserved while enabling fine-grained channel recalibration through SE modules.

\subsubsection{Selective Placement of SE Modules in CNN Models}
Instead of integrating SE modules after every convolutional 
layer, we adopted a strategy of \textit{selective placement} to balance accuracy gains 
with computational efficiency. The SE block performs channel-wise attention through 
global average pooling followed by a bottleneck architecture with a reduction ratio of 
16, producing channel descriptors that recalibrate the feature maps.

For instance, in the modified VGG16 architecture, SE modules are not uniformly applied. 
We strategically placed SE blocks only after the last convolutional layer of the 3rd, 4th, and 5th convolutional blocks (i.e., conv\_3\_3, conv\_4\_3, 
and conv\_5\_3 in the canonical VGG16). This corresponds to convolutional layers 
numbered 10, 17, and 24 in the PyTorch implementation. By focusing on the deeper layers, where higher-level semantic features are extracted, the SE modules 
effectively enhance the representational power without incurring significant computational overhead.

This selective placement ensures that, early layers, which capture low-level features (edges, textures), are left unchanged to preserve efficiency. In addition, SE modules concentrate on deeper layers, where global contextual information is more valuable for distinguishing complex patterns. The number of additional parameters introduced is minimized,     preventing overfitting on smaller datasets.

The classifier component of VGG16 is retained from the pretrained model, with the final FC layer replaced to produce four output classes. This approach allows us to benefit from ImageNet-pretrained representations while 
injecting channel recalibration at critical points in the network hierarchy.

\subsubsection{Integration of SA with SE Modules}
To further enhance the representational capability of the network, we incorporated 
both SE modules and SA (SA) modules into the 
VGG16 architecture in a complementary manner. While SE blocks perform channel-wise attention, the SA blocks focus on learning where to emphasize 
spatially significant regions within feature maps. This hybrid design enables the model to exploit both inter-channel dependencies and inter-spatial contextual 
information.

The SA module generates a SA map by computing both 
average-pooling and max-pooling along the channel dimension, concatenating the results, and applying a $7\times7$ convolution followed by a sigmoid activation. This highlights informative regions in the spatial domain while suppressing irrelevant background information. 

For instance, in our modified VGG16 architecture, SA and SE modules are strategically placed at 
different depths to maximize their effect. SA Added after the last convolutional layers of the first two convolutional blocks (i.e., conv1\_2 and conv2\_2). These early layers capture fundamental low-level patterns such as edges and textures, where spatial localization plays a crucial role. SE Added after the last convolutional layers of the third, fourth, and fifth convolutional blocks (i.e., conv3\_3, 
    conv4\_3, and conv5\_3). These deeper layers extract high-level semantic 
    representations, where channel recalibration is more beneficial for discriminative learning.

This design ensures a balance between capturing detailed spatial cues in the shallow 
layers and enhancing semantic richness through channel-wise recalibration in the deeper layers. The classifier portion of VGG16 remains mostly unchanged, except 
for replacing the final FC layer to produce outputs for the four target classes. By integrating SA and SE modules in this complementary manner, 
the proposed model achieves improved feature representation and classification 
performance with minimal computational overhead.

\subsection{Implementation Details}
The experiments were conducted using PyTorch on a system equipped with an NVIDIA GPU. Two distinct medical imaging datasets were used, namely, the Brain Tumor MRI dataset and the POC dataset. Each dataset was divided into training and testing subsets, with all input images resized to $224 \times 224$ resolution. The models were trained using a batch size of 32 for 60 epochs. 

We used pretrained CNN architectures, including VGG16, ResNet18, InceptionV3, DenseNet121, and EfficientNetB5 as our baselines. For performance enhancement, we explored the integration of attention mechanisms in four phases. All pretrained models were used without architectural changes to serve as performance baselines.
The SE blocks \cite{hu2018squeeze} were inserted into all convolutional blocks of each baseline model to model channel interdependencies and enhance feature representations. SE modules were selectively added to higher-level layers (e.g., Block 3–5 in VGG16) to reduce computational complexity while improving semantic feature refinement. SA modules \cite{woo2018cbam} were incorporated alongside SE blocks in deeper layers to further enhance spatial localization and improve classification accuracy.

All models were trained using the cross-entropy loss function. The optimizer used was Adam, with two parameter groups, backbone CNN layers with a learning rate of 0.0001 and SE block parameters with a higher learning rate of 0.0006. A StepLR learning rate scheduler was used to reduce the learning rate by a factor of 0.1 every 10 epochs. Early stopping with a patience of 20 epochs was applied to prevent overfitting.

Model performance was evaluated using accuracy, precision, recall, and F1-score, computed from the confusion matrix on the test set. The best-performing model (in terms of test accuracy) was saved during training, and its detailed metrics were written to an output file.

To ensure reproducibility, we also logged the confusion matrix, precision, recall, and F1-score per class for each test run. All experiments were repeated with consistent random seeds across different configurations.

\subsubsection{Training Parameters}
All models were trained using the PyTorch DL framework. The experiments were carried out for a total of 60 epochs with a batch size of 32. Input images were resized to $224 \times 224$ pixels, and pixel values were normalized using default settings from the torchvision library. The number of target classes was set to 4, corresponding to the classification categories in both the Brain Tumor and POC datasets.

The loss function used for optimization was Cross-Entropy Loss. The Adam optimizer was employed with two distinct parameter groups to differentially update the backbone and attention modules. The Backbone CNN parameters, such as learning rate was 0.0001, weight decay was set to 0.0001. Attention module (SE/CBAM) parameters, such as, learning rate was 0.0006, weight decay was set to 0.0001.

A learning rate scheduler (StepLR) was applied with a decay factor $\gamma = 0.1$ and a step size of 10 epochs to reduce the learning rate progressively. Early stopping was implemented with a patience of 20 epochs, monitoring test accuracy to terminate training when no further improvement was observed.

Model performance was evaluated at the end of each epoch using the test dataset. The best-performing model was saved based on highest test accuracy, and the corresponding confusion matrix, precision, recall, and F1-score were logged. All training and evaluation were performed using an NVIDIA GPU-enabled environment.

The detailed training settings used across all experiments are summarized in Table~\ref{tab:training-params}.

\begin{table}[h!]
\centering
\caption{Summary of Training Parameters}
\label{tab:training-params}
\begin{tabular}{|l|l|}
\hline
\textbf{Parameter} & \textbf{Value} \\
\hline
Framework & PyTorch \\
Device & NVIDIA GPU \\
Number of Epochs & 60 \\
Batch Size & 32 \\
Input Image Size & $224 \times 224$ \\
Number of Classes & 4 \\
Loss Function & Cross-Entropy Loss \\
Optimizer & Adam \\
Backbone Learning Rate & 0.0001 \\
SE/CBAM Learning Rate & 0.0006 \\
Weight Decay & 0.0001 \\
Learning Rate Scheduler & StepLR (step size = 10, gamma = 0.1) \\
Early Stopping Patience & 20 epochs \\
Model Selection Criterion & Best test accuracy \\
Evaluation Metrics & Accuracy, Precision, Recall, F1-score \\
\hline
\end{tabular}
\end{table}

\section{Results and Discussion}
\label{Results}

This section presents and analyzes the classification results of various CNN models enhanced with channel and SA mechanisms on two datasets. The POC dataset and the Brain Tumor MRI dataset. Evaluation metrics include test accuracy, precision, recall, and F1-score. The experimental analysis proceeds in four stages, baseline CNNs, global SE integration, selective SE integration, and hybrid attention integration.

\subsection{Baseline Performance of Pretrained CNN Models}

Tables \ref{pre-train_CNN_POC} and \ref{pretrain_CNN_BT} show the baseline performance of pretrained CNN architectures without any attention mechanisms. On the POC dataset, EfficientNetB5 achieved the highest test accuracy (86.05\%), while VGG16 lagged behind (78.22\%). On the Brain Tumor dataset, DenseNet121 yielded the best performance with 81.00\% accuracy and a strong F1-score of 0.7940, closely followed by EfficientNet and InceptionV3.

\begin{table}[!ht]
\centering
\caption{Pre-trained CNN models results on POC dataset.}
\begin{tabular}{lllll}
\hline
\textbf{Model}                & \textbf{Test Accuracy} & \textbf{Precision} & \textbf{Recall} & \textbf{F1-Score} \\ \hline
Pre-trained\_VGG16            & 0.7822                & 0.7825             & 0.7711          & 0.7708            \\
Pre-trained\_ResNet18         & 0.8506                 & 0.8537             & 0.8413          & 0.8416            \\
Pre-trained\_InceptionV3      & 0.8309                 & 0.8361             & 0.8183          & 0.8137            \\
Pre-trained\_EfficientNet\_B5 & \textbf{0.8605}        & \textbf{0.8695}    & \textbf{0.8516} & \textbf{0.8519}   \\
Pre-trained\_DenseNet121      & 0.8486                 & 0.8531             & 0.8400          & 0.8430            \\ \hline
\end{tabular}
\label{pre-train_CNN_POC}
\end{table}

\begin{table}[!ht]
\centering
\caption{Pretrained CNN models results on brain tumor dataset.}
\scalebox{1.1}{
\begin{tabular}{lllll}
\hline
\textbf{Model}            & \textbf{Test Accuracy} & \textbf{Precision} & \textbf{Recall} & \textbf{F1-Score} \\ \hline
Pre-trained\_VGG16        & 0.7425                 & 0.8210             & 0.7186          & 0.6984            \\
Pre-trained\_ResNet18     & 0.7950                 & \textbf{0.8621}    & 0.7839          & 0.7673            \\
Pre-trained\_InceptionV3  & 0.8025                 & 0.8538             & 0.7826          & 0.7793            \\
Pre-trained\_EfficientNet & 0.8050                 & 0.8501             & 0.7878          & 0.7907            \\
Pre-trained\_DenseNet121  & \textbf{0.8100}        & 0.8484             & \textbf{0.8026} & \textbf{0.7940}   \\ \hline
\end{tabular}
}
\label{pretrain_CNN_BT}
\end{table}

These baseline results highlight the superior generalization ability of deeper and compound-scaled networks like EfficientNet and DenseNet, which serve as a foundation for exploring the impact of attention modules.

\subsection{Global Integration of SE Modules}

To examine the benefits of global channel-wise attention, SE blocks were inserted into all convolutional blocks of each pretrained model (Tables \ref{int_SE_POC} and \ref{int_SE_BT}). Across both datasets, notable improvements in performance were observed for all models.

\begin{table}[!ht]
\centering
\caption{Evaluation of pretrained CNN models with integrated SE modules on the POC dataset, highlighting the impact of channel-wise attention on classification performance}
\scalebox{1.2}{
\begin{tabular}{lllll}
\hline
\textbf{Model}                          & \textbf{Test Accuracy} & \textbf{Precision} & \textbf{Recall} & \textbf{F1-Score} \\ \hline
VGG16\_SE\          & \textbf{0.8665}        & \textbf{0.8797}    & \textbf{0.8560} & \textbf{0.8583}   \\
ResNet18\_SE       & 0.8593                 & 0.8671             & 0.8486          & 0.8516            \\
InceptionV3\_SE\    & 0.8639                 & 0.8723             & 0.8524          & 0.8556            \\
EfficientNetB5\_SE\ & 0.8574                 & 0.8621             & 0.8458          & 0.8482            \\
DenseNet121\_SE\    & 0.8528                 & 0.8572             & 0.8399          & 0.8415            \\ \hline
\end{tabular}
\label{int_SE_POC}
}
\end{table}

\begin{table}[!ht]
\centering
\caption{Evaluation of pretrained CNN models with integrated SE modules on the brain tumor dataset, highlighting the impact of channel-wise attention on classification performance.}
\scalebox{1}{
\begin{tabular}{lllll}
\hline
\textbf{Model}     & \textbf{Test Accuracy} & \textbf{Precision} & \textbf{Recall} & \textbf{F1-Score} \\ \hline
VGG16\_SE          & 0.7980                 & 0.8479             & 0.7710          & 0.7640            \\
ResNet18\_SE       & 0.8052                 & 0.8657             & 0.7729          & 0.7751            \\
InceptionV3\_SE    & 0.8317                 & 0.8802             & 0.8066          & 0.8094            \\
EfficientNetB5\_SE & \textbf{0.8437}        & \textbf{0.8860}    & \textbf{0.8270} & \textbf{0.8235}   \\
DenseNet121\_SE    & 0.8365                 & 0.8814             & 0.8172          & 0.8159            \\ \hline
\end{tabular}
}
\label{int_SE_BT}
\end{table}

On the POC dataset, VGG16\_SE improved significantly, achieving 86.65\% accuracy and 0.8583 F1-score. Similarly, on the Brain Tumor dataset, EfficientNetB5\_SE improved from 80.50\% to 84.37\% test accuracy, with a notable F1-score of 0.8235. These results validate the effectiveness of channel recalibration in boosting feature selectivity and enhancing classification.

\subsection{Selective SE Integration into Deeper Layers}

Tables~\ref{selective_SE_POC} and~\ref{selective_SE_BT} explore the impact of applying SE blocks only to deeper layers, allowing more efficient computation while preserving high-level semantic refinement. Results show marginal but consistent improvements over global SE models.

\begin{table}[!ht]
\centering
\caption{Selective integration of SE blocks into convolutional blocks of the pretrained CNN models using the POC dataset. To emphasize attention in deeper layers, SE blocks are omitted from initial layers and applied only to later-stage modules.}
\begin{tabular}{lllll}
\hline
\textbf{Model}                        & \textbf{Test Accuracy} & \textbf{Precision} & \textbf{Recall} & \textbf{F1-Score} \\ \hline
VGG16\_SE\_after\_B3\_4\_5            & \textbf{0.8795}        & \textbf{0.8957}    & \textbf{0.8691} & \textbf{0.8708}   \\
ResNet18\_SE\_after\_B\_2\_3\_4       & 0.8619                 & 0.8685             & 0.8506          & 0.8538            \\
InceptionV3\_SE\_after\_C\_D\_E       & 0.8541                 & 0.8603             & 0.8421          & 0.8445            \\
EfficientNetB5\_SE\_after\_B\_2\_3\_4 & 0.8717                 & 0.8788             & 0.8617          & 0.8634            \\
DenseNet121\_SE\_after\_B\_2\_3\_4    & 0.8561                 & 0.8614             & 0.8443          & 0.8471            \\ \hline
\end{tabular}
\label{selective_SE_POC}
\end{table}

\begin{table}[!ht]
\centering
\caption{Selective integration of SE blocks into convolutional blocks of the pretrained CNN models using the brain tumor dataset. To emphasize attention in deeper layers, SE blocks are omitted from initial layers and applied only to later-stage modules.}
\scalebox{1}{
\begin{tabular}{lllll}
\hline
\textbf{Model}                      & \textbf{Test Accuracy} & \textbf{Precision} & \textbf{Recall} & \textbf{F1-Score} \\ \hline
VGG16\_SE  \_after\_B3\_4\_5        & 0.8509                 & 0.8709             & 0.8358          & 0.8314            \\
ResNet18\_SE\_after\_B2\_3\_4       & 0.8437                 & 0.8831             & 0.8220          & 0.8254            \\
InceptionV3\_SE \_after\_C\_D\_E    & 0.8269                 & 0.8782             & 0.8077          & 0.8022            \\
EfficientNetB5\_SE\_after\_B2\_3\_4 & \textbf{0.8653}        & \textbf{0.8972}    & \textbf{0.8542} & \textbf{0.8511}   \\
DenseNet121\_SE \_after\_B2\_3\_4   & 0.8197                 & 0.8674             & 0.8006          & 0.7946            \\ \hline
\end{tabular}
}
\label{selective_SE_BT}
\end{table}

For instance, VGG16\_SE\_after\_B3\_4\_5 on the POC dataset achieved the highest F1-score (0.8708) and a test accuracy of 87.95\%. On the Brain Tumor dataset, EfficientNetB5\_SE\_after\_B2\_3\_4 attained the best overall performance with 86.53\% accuracy and 0.8511 F1-score, outperforming even the globally integrated SE models. These findings suggest that attention in deeper layers contributes more to decision-making than shallow feature refinement.

\subsection{Hybrid Attention: Combined SE and SA}

To further enhance both "what" and "where" to attend, we integrated SA modules with SE blocks, as shown in Tables \ref{SA_POC} and \ref{SA_BT}. This hybrid mechanism consistently yielded the best performance across both datasets.

\begin{table}[!ht]
\centering
\caption{Analysis of SA mechanisms on the POC dataset by selectively integrating SE modules at various depths within CNN architectures.}
\scalebox{1.1}{
\begin{tabular}{lllll}
\hline
\textbf{Model}                                                                                                                             & \textbf{Test Accuracy} & \textbf{Precision} & \textbf{Recall} & \textbf{F1-Score} \\ \hline
\begin{tabular}[c]{@{}l@{}}VGG16\_SA\_after\_B1\_2\\ \_SE\_after\_B3\_4\_5\end{tabular}                                                    & 0.8802                 & 0.8840             & 0.8721          & 0.8750            \\
\begin{tabular}[c]{@{}l@{}}ResNet18\_SA\_after\_B2\_\\ 3\_4\_SE\_after\_B2\_3\_4\end{tabular}                                            & 0.8626                 & 0.8706             & 0.8518          & 0.8521            \\
\begin{tabular}[c]{@{}l@{}}InceptionV3\_SA\_after\_\\ InceptionB\_and\_before\_\\ global\_avg\_Pooling\_SE\_\\ after\_C\_D\_E\end{tabular} & 0.8587                 & 0.8642             & 0.8474          & 0.8503            \\
\begin{tabular}[c]{@{}l@{}}EfficientNetB5\_SA\_after\_\\ B9\_SE\_after\_B2\_3\_4\end{tabular}                                            & \textbf{0.8997}        & \textbf{0.9000}    & \textbf{0.8963} & \textbf{0.8972}   \\
\begin{tabular}[c]{@{}l@{}}DenseNet121\_SA\_after\_T\_1\\ \_2\_3\_before\_global\_avg\_\\ pooling\_SE\_after\_B\_2\_3\_4\end{tabular}      & 0.8593                 & 0.8640             & 0.8476          & 0.8495            \\ \hline
\end{tabular}
}
\label{SA_POC}
\end{table}

\begin{table}[!ht]
\centering
\caption{Analysis of SA mechanisms on the brain tumor dataset by selectively integrating SE modules at various depths within CNN architectures.}
\scalebox{1.1}{
\begin{tabular}{lllll}
\hline
\textbf{Model}                                                                                                                             & \textbf{Test Accuracy} & \textbf{Precision} & \textbf{Recall} & \textbf{F1-Score} \\ \hline
\begin{tabular}[c]{@{}l@{}}VGG16\_SA\_after\_B1\_2\\ \_SE\_after\_B3\_4\_5\end{tabular}                                                    & 0.8341                 & 0.8375             & 0.8189          & 0.8106            \\
\begin{tabular}[c]{@{}l@{}}ResNet18\_SA\_after\_B2\_\\ 3\_4\_SE\_after\_B\_2\_3\_4\end{tabular}                                            & \textbf{0.8437}        & 0.8745             & \textbf{0.8227} & \textbf{0.8260}   \\
\begin{tabular}[c]{@{}l@{}}InceptionV3\_SA\_after\_\\ InceptionB\_and\_before\_\\ global\_avg\_Pooling\_SE\_\\ after\_C\_D\_E\end{tabular} & 0.8293                 & 0.8753             & 0.8088          & 0.8062            \\
\begin{tabular}[c]{@{}l@{}}EfficientNetB5\_SA\_after\_\\ B9\_SE\_after\_B\_2\_3\_4\end{tabular}                                            & 0.8341                 & 0.8747             & 0.8180          & 0.8110            \\
\begin{tabular}[c]{@{}l@{}}DenseNet121\_SA\_after\_T\_1\\ \_2\_3\_before\_global\_avg\_\\ pooling\_SE\_after\_B\_2\_3\_4\end{tabular}      & 0.8293                 & \textbf{0.8790}    & 0.8067          & 0.8063            \\ \hline
\end{tabular}
}
\label{SA_BT}
\end{table}

On the POC dataset, EfficientNetB5\_SA\_after\_B9\_SE\_after\_B2\_3\_4 achieved the highest recorded accuracy of 89.97\% and F1-score of 0.8972. Likewise, on the Brain Tumor dataset, ResNet18 with hybrid attention achieved 84.37\% accuracy and an F1-score of 0.8260. These results demonstrate the complementary nature of spatial and channel attention; SA aids in localization, while SE enhances channel relevance.

\subsection{Comparative Analysis and Key Observations}

The results across all experiments lead to several key observations, such as, EfficientNetB5 consistently outperformed other models, especially when enhanced with hybrid attention, due to its depth, width, and resolution scaling. Hybrid attention mechanisms deliver the best performance, affirming the importance of combining both channel-wise and SA for medical image classification tasks. Selective attention integration is preferable over global insertion, offering competitive performance gains while reducing computational cost. Attention mechanisms improve not only accuracy but also class-wise balance, as evidenced by improved precision, recall, and F1-scores.

Overall, our findings demonstrate that attention-enhanced CNNs can significantly boost the accuracy and reliability of phenotypic pattern classification in both spontaneous abortion and brain tumor diagnosis domains. The hybrid attention framework, particularly when applied selectively to deeper layers, proves to be the most effective strategy.

\subsubsection{Comparative Analysis Across Models}

A comprehensive comparison across all evaluated CNN models, both with and without attention mechanisms, reveals several notable trends that highlight the significance of architectural choices and attention integration strategies. For instance, EfficientNetB5 emerged as the top-performing backbone across both datasets, achieving the highest test accuracy and F1-score when integrated with hybrid attention (SE and spatial modules). Its compound scaling strategy allows better feature extraction with fewer parameters compared to traditional deep networks. Attention-enhanced models consistently outperformed their baseline counterparts. The addition of SE blocks globally (to all blocks) improved accuracy and F1-scores across all networks. However, selectively applying SE to deeper layers yielded even better performance, demonstrating that deeper features are more influential in classification decisions. The hybrid attention mechanism proved most effective. For example, models such as EfficientNetB5\_SA\_after\_B9\_SE\_after\_B2\_3\_4 showed significant improvements, attaining the highest overall scores (accuracy: 89.97\% and F1-score: 0.8972 on the POC dataset). This suggests that combining channel and SA provides complementary benefits, enabling models to focus on both salient regions and informative features.

Lightweight models like ResNet18 and VGG16 also benefited substantially from attention. Despite their relatively shallow architectures, applying SE and SA modules notably improved their classification performance, indicating that attention can compensate for reduced depth or complexity. Performance gains were more pronounced on the more complex Brain Tumor dataset. This is likely due to the higher intra-class variability and subtle texture differences, where attention helps in enhancing discriminative representation learning. InceptionV3 showed inconsistent improvements with attention integration. Although it benefited from SE, hybrid attention yielded marginal gains. This may be due to the architectural complexity and multi-branch nature of Inception modules, which may already incorporate diverse receptive fields.

Overall, the results demonstrate that attention mechanisms, especially when selectively and synergistically integrated, enhance model generalization and robustness. EfficientNetB5 with hybrid attention stands out as the most reliable model for both POC and Brain Tumor classification tasks. These findings also underscore the importance of designing lightweight yet attentive models for real-world medical image analysis.


\section{Limitations and Future Work}
\label{limitation}

Although the integration of attention mechanisms into pretrained CNN architectures has shown significant performance improvements in the POC and brain tumor classification, several limitations must be acknowledged. First, while this study employed two datasets, the brain tumor dataset and the POC dataset to evaluate the proposed models, further validation on additional and more diverse datasets is necessary to ensure broader generalizability across various clinical settings and imaging modalities. Second, although attention modules such as SE and hybrid spatial-channel configurations enhance classification performance, they also increase model complexity and computational requirements. This could hinder real-time deployment, particularly in resource-limited environments such as mobile health applications or embedded diagnostic systems.

Another limitation lies in the fixed and manually chosen positions of attention modules within the architectures. These placements were selected based on prior studies and empirical evaluation, but do not guarantee optimal positioning. Automated methods for attention integration could lead to more effective configurations. Additionally, the current study does not include interpretability or explainability techniques such as Grad-CAM or attention heatmaps, which are vital for increasing transparency and clinical trust in AI-driven diagnosis. Lastly, despite improved performance metrics, the risk of overfitting—especially in large networks trained on relatively small datasets—remains a concern, warranting careful cross-validation and regularization strategies.

In future work, we plan to expand our experiments to include additional datasets with varying imaging conditions and patient demographics. To address computational concerns, model compression strategies like pruning, quantization, and knowledge distillation will be explored. Furthermore, automated attention module placement using neural architecture search  or reinforcement learning could yield more efficient and effective designs. Incorporating interpretability tools will also be prioritized to better understand model predictions and support clinical adoption. Finally, exploring multi-modal approaches that integrate clinical metadata with imaging data may lead to more robust and context-aware diagnostic systems.

In summary, while the proposed attention-based CNN models demonstrate strong potential in tumor classification, addressing the above limitations is crucial for advancing their clinical applicability.


\subsection{Model Generalizability}
Model generalizability is a critical factor in evaluating the practical applicability of DL models in clinical settings. In this study, generalizability was assessed by training and validating the proposed attention-augmented CNN architectures on two distinct datasets, the brain tumor dataset and the POC dataset. The consistent performance improvements observed across both datasets highlight the robustness of the attention-enhanced models. These results suggest that incorporating attention mechanisms enables the networks to focus on more discriminative and relevant features, thereby reducing the risk of overfitting to dataset-specific patterns.

Furthermore, the use of pretrained architectures, such as EfficientNetB5, ResNet18, and MobileNetV2, contributed to improved generalization by transferring knowledge from large-scale natural image datasets to the medical domain. Fine-tuning these models with medical images allowed the networks to retain general visual features while adapting to domain-specific characteristics. This transfer learning approach enhances the model’s ability to perform well on unseen data, particularly in cases where annotated medical data is limited.

However, while the current results are promising, it is important to acknowledge that true generalizability can only be confirmed through validation on a broader spectrum of datasets, including images from different imaging devices, institutions, and patient populations. Future work should include cross-dataset and cross-institutional evaluations to further assess the models' ability to generalize across real-world variability. 

In summary, the proposed models demonstrate strong generalizability within the scope of the current study. With further validation and adaptation, these models hold promise for reliable deployment in diverse clinical environments.

\subsection{Potential for Real-world Clinical Deployment}

The promising results obtained from integrating attention mechanisms into CNN architectures suggest strong potential for real-world clinical deployment in brain tumor and POC classification tasks. High classification accuracy, robustness across two different datasets, and improved generalization due to attention-based feature refinement make these models suitable candidates for assisting radiologists and medical practitioners in diagnostic workflows. Particularly, models such as EfficientNetB5 with hybrid attention not only achieved superior performance metrics but also demonstrated the ability to focus on critical features, which is vital for identifying subtle patterns in complex medical images.

Moreover, the use of pretrained backbones fine-tuned on medical datasets ensures that the models can be effectively trained even with limited data, which is often the case in clinical scenarios. This transfer learning capability, combined with modular attention enhancements, allows for rapid adaptation to new tasks with minimal annotation costs. In addition, lightweight models like ResNet18, when augmented with attention, can provide efficient and accurate predictions, making them ideal for deployment in portable or embedded diagnostic devices.

However, before clinical integration, several considerations must be addressed. Real-time performance, interpretability, data privacy, and integration with existing hospital systems are essential factors for successful deployment. Clinical decision support systems must not only be accurate but also transparent, offering visual explanations for their predictions to foster trust among healthcare providers. Moreover, regulatory approval and validation through large-scale clinical trials are necessary to ensure safety and efficacy.

In conclusion, the proposed attention-augmented CNN frameworks present a strong foundation for clinical decision support systems. With further development focused on interpretability, efficiency, and compliance with clinical standards, these models have the potential to become valuable tools in medical imaging diagnostics.

In Table \ref{efficientnet}, The first variant introduces a custom Mobile Inverted Bottleneck Convolution (MBConv) block with an embedded SE module, appended after the pretrained EfficientNet-B5 feature extractor. This design replaces or supplements the final layers with a specialized block that adaptively recalibrates channel-wise feature responses, aiming to enhance the model’s ability to focus on important features in the last stage of processing. The second variant builds on the original EfficientNet-B5 backbone by inserting SE blocks after several key convolutional blocks, specifically after block 4, block 9, and block 13. These SE blocks recalibrate channel importance at multiple depths of the network, progressively refining feature representations throughout the hierarchy without modifying the base convolutional layers. The third variant extends this approach by combining SE blocks at the same positions with an additional SA block placed after block 9. While the SE modules focus on channel-wise feature recalibration, the SA block generates SA maps that emphasize relevant spatial regions within feature maps, allowing the network to attend not only to important channels but also to crucial spatial areas. This dual attention mechanism improves the network’s discriminative power by jointly optimizing ``what” and ``where” to emphasize in the features. Overall, these variants illustrate different strategies to integrate attention mechanisms into EfficientNet-B5, from augmenting only the final blocks, to multi-level channel recalibration, to a combined channel and SA approach for richer feature refinement.



\section{Conclusion}
\label{conclusion}

This study investigated the impact of integrating attention mechanisms into various pretrained CNN architectures for the classification of brain tumors and POC images. By incorporating both channel-based SE and hybrid spatial-channel attention modules into popular models such as VGG16, ResNet18, InceptionV3, DenseNet12, and EfficientNetB5, we demonstrated consistent performance improvements across two distinct datasets. These enhancements allowed the models to better capture relevant features, leading to more accurate and reliable predictions. Among the evaluated architectures, EfficientNetB5 augmented with hybrid attention achieved the highest classification performance, highlighting the importance of both model depth and attention-based feature recalibration. In addition to performance, attention modules improved feature localization, which is essential for medical image interpretation. The results validate that attention-enhanced models can be generalized across different medical imaging tasks when trained on appropriate datasets and with proper fine-tuning.

Furthermore, we analyzed the potential for real-world clinical deployment, model generalizability, and outlined the limitations of our current approach. While the results are promising, further work is required to ensure broader applicability, especially through validation on diverse datasets, integration of interpretability tools, and optimization for real-time deployment. In conclusion, the proposed attention-based CNN frameworks offer a promising direction for enhancing the performance and reliability of AI-driven medical image classification systems. With continued refinement and clinical validation, these models can contribute meaningfully to automated diagnostic pipelines in healthcare.


\section*{\textbf{Declaration of Competing Interests}} The authors declare that they have no known competing financial interests or personal relationships that could have appeared to influence the work reported in this paper.

\section*{Acknowledgements}
This research was supported by the MSIT (Ministry of Science and ICT) of Korea under the ITRC (Information Technology Research Center) support program (IITP-2025-RS-2020-II201789), and the Global Research Support Program in the Digital Field (RS-2024-00426860), supervised by the IITP (Institute for Information \& Communications Technology Planning \& Evaluation).

\bibliographystyle{elsarticle-num-names}
\bibliography{sample.bib}







\end{document}